\documentclass[letterpaper, 10 pt, conference]{ieeeconf}
\usepackage[utf8]{inputenc}
\usepackage{cite}
\usepackage{graphicx}
\graphicspath{{figures/}}
\usepackage{amsmath,amssymb,amsfonts,bbm}
\usepackage{caption, subcaption}
\usepackage{algorithmic}
\usepackage{textcomp}
\usepackage{soul}
\usepackage{xcolor}
\usepackage{acronym}

\acrodef{pHRI}[p\textsc{HRI}]{physical human-robot interaction}
\acrodef{DOF}[\textsc{DOF}]{degrees of freedom}
\acrodef{DMP}[\textsc{DMP}]{Dynamic Movement Primitive}
\acrodef{HZD}[\textsc{HZD}]{Hybrid Zero Dynamics}
\acrodef{ZMP}[\textsc{ZMP}]{Zero Moment Point}
\acrodef{NN}[\textsc{NN}]{neural network}
\acrodef{PAC}[\textsc{PAC}]{Probably Approximately Correct}
\acrodef{iid}[i.i.d.]{independent and identically distributed}
\acrodef{ES}[\textsc{ES}]{Evolutionary Strategies}

\newtheorem{theorem}{\bf Theorem}
\newtheorem{remark}{\bf Remark}

\def\BibTeX{{\rm B\kern-.05em{\sc i\kern-.025em b}\kern-.08em
    T\kern-.1667em\lower.7ex\hbox{E}\kern-.125emX}}

\begin{document}

\title{\textbf{Interactive Dynamic Walking: \\ Learning Gait Switching Policies with Generalization Guarantees}}

\author{Prem Chand, Sushant Veer, and Ioannis Poulakakis
\thanks{P. Chand and I. Poulakakis are with the Department of Mechanical Engineering, University of Delaware, Newark, DE 19716, USA;  e-mail: {\tt\small \{premc, poulakas\}@udel.edu.} S. Veer is with the Department of Mechanical and Aerospace Engineering, Princeton University, Princeton, NJ 08544, USA; e-mail: {\tt\small sveer@princeton.edu.}}
\thanks{This work was supported in part by NSF CAREER Award IIS-1350721.}
}

\maketitle

\pagestyle{empty}
\thispagestyle{empty}

\begin{abstract}
In this paper, we consider the problem of adapting a dynamically walking bipedal robot to follow a leading co-worker while engaging in tasks that require physical interaction. Our approach relies on switching among a family of Dynamic Movement Primitives (DMPs) as governed by a supervisor. We train the supervisor to orchestrate the switching among the DMPs in order to adapt to the leader's intentions, which are only \emph{implicitly} available in the form of interaction forces. The primary contribution of our approach is its ability to furnish \emph{certificates of generalization} to novel leader intentions for the trained supervisor. This is achieved by leveraging the Probably Approximately Correct (PAC)-Bayes bounds from generalization theory. We demonstrate the efficacy of our approach by training a neural-network supervisor to adapt the gait of a dynamically walking biped to a leading collaborator whose intended trajectory is not known explicitly.
\end{abstract}

\section{Introduction}

Imagine a bipedal robot physically assisting a human to carry an object. The human knows where and how the object needs to be placed and the spatial layout of the area. Based on this information, a plan of action can be quickly devised by the human. More often than not, however, this plan cannot be explicitly---e.g., as a desired trajectory---communicated to the robot. Yet, in physically coupled dyads, the forces developed at the port of interaction between the robot and the human encode important information regarding the human's action plan. This paper aims at enabling bipedal robots to make decisions as to how \emph{to interact} and modify their gaits to follow intended---yet \emph{unknown}---trajectories.

\subsection{Related Work}

Being able to adapt to intentions communicated \emph{indirectly} via interaction forces is a central problem in \ac{pHRI}. In this context, admittance control has been widely employed to ``translate'' externally applied forces to motion control references; see~\cite{Keemink2018IJRR} for a recent overview. The choice of the parameters that govern the dynamic relation between the force and the desired motion is critical in such tasks~\cite{Ikeura1994IWRHC, Ikeura1995ICRA}, and a variety of adaptive admittance control schemes has been proposed to enhance transparency. Although, a detailed account of these schemes would take us too far afield, it is of interest here to briefly describe how ``intention'' is estimated and then used to adapt the robot's behavior accordingly. 

To enhance the robot's responsiveness, early work in~\cite{Ikeura1995ICRA} developed a simple velocity-based switching admittance rule that deduces the collaborator's intention by comparing the velocity with a given threshold. However, estimating this threshold requires a priori acquisition of a typical movement for the task at hand. To avoid this restriction and increase transparency,~\cite{Duchaine2007WHC} proposed the rate of change of the external force as an indicator for the intended motion---accelerate or decelerate---while~\cite{Bae2020Access} used force-based calculations of the desired acceleration and velocity to provide better discrimination among the collaborator's intentions. 
However, these methods assume only a few intention states and are restricted to simple point-to-point collaborative tasks.
To address more general situations,~\cite{Li2014TMECH} represented the collaborator's intentions by a desired trajectory which is \emph{not} explicitly known and is estimated online via a \ac{NN}. Importantly, in this work, the estimated position becomes the rest position of an adaptive impedance controller so that the robot can \emph{actively} follow the collaborator's intended trajectory without acting as a load to its partner~\cite{Li2014TMECH}.
Another approach is discussed in~\cite{ranatunga2015intent}, which integrates a neuroadaptive controller with an outer loop that uses human walking path predictions to convert applied forces to desired position.

In this paper, we address the challenge of translating forces representing intention to desired motions through a learning-based approach. However, unlike previously proposed methods, our approach is accompanied with provable performance guarantees on novel collaborator intentions; that is, \emph{generalization guarantees}.
Generalization refers to the ability of a learned function to perform well on test sets that are different from the training sets, albeit drawn from the same distribution. 
%
%
Our approach falls within the purview of the \ac{PAC}-Bayes framework, which recently demonstrated the ability to provide strong generalization guarantees on deep neural networks~\cite{rivasplata2019pac}. Harnessing the recently developed \ac{PAC}-Bayes control framework~\cite{majumdar2020pac,veer2020probably}, we provide a method for learning policies accompanied with explicit bounds on performance under novel leader intentions, communicated via interaction forces.
We believe that bounds of this sort are important in the context of \ac{pHRI}, since they essentially quantify the \emph{risk} of applying control policies to robots engaged in tasks that involve physical contact.

We apply the proposed learning-based approach to the case of a bipedal robot walking under the influence of an interaction force corresponding to the intended trajectory of a leading collaborator. This scenario is representative of a class of tasks in which a human and a biped physically collaborate to transport an object. 
Beyond estimating the collaborator's intention, engaging legged robots in such tasks presents an additional challenge: locomotion stability. The majority of existing methods---see~\cite{Ervard2012SR, Bussy2012IROS, Berger2013HUMAN} for example---ensure walking stability via the \ac{ZMP} criterion, which can be suitably combined with reactive walking pattern generators to ensure that the robot safely adapts to the collaborator's intentions.
Unlike \ac{ZMP}-based walkers, dynamically walking bipeds~\cite{mcgeer1990passive} have not been studied in the context of tasks that involve interaction. This is because combining walking stability with interaction in such systems is challenging. As a result, controller development for dynamically walking bipeds has been restricted to locomotion stability alone~\cite{westervelt2007feedback, Reher2020ACC, Castillo2021Arxiv}, with recent results also addressing motion planning~\cite{motahar2016composing, veer2017driftless, Teng2021arXiv} and trajectory following~\cite{Xiong2020Arxiv} under the assumption that the plan and the desired trajectory are known. 
In this work, we develop a hierarchically structured algorithm that takes advantage of existing locomotion control methods to bring dynamic walkers a step closer to executing tasks that involve following unknown intended trajectories based on interaction forces.

%

\subsection{Overview: Adaptation via PAC-Bayes Switching}

\begin{figure}[b!]
\vspace{-0.25in}
\begin{center}
\includegraphics[width=0.95\columnwidth]{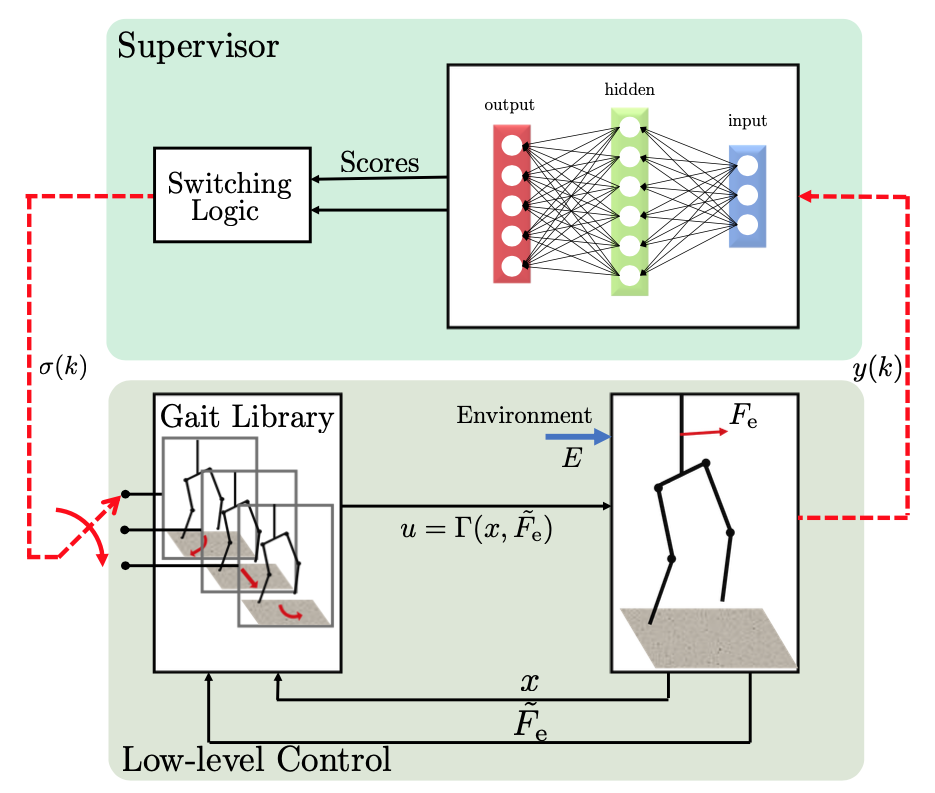}
\end{center}
\vspace{-0.1in}
\caption{Overview of the proposed framework. The high-level supervisor processes stride-to-stride observations $y(k)$ about the robot's motion and interaction forces, and returns the index $\sigma(k)$ of the gait controller that must be engaged during the $k$-th step. The supervisor relies on a \ac{NN} that learns---with provable generalization guarantees---how to interpret interaction force measurements as a directional indicator of the unknown desired trajectory. The low-level feedback loop executes the suggested gait controller.}
\label{fig:block_diagram} 
\end{figure}

Building on our recent results~\cite{veer2017supervisory, Veer2019ICRA, Veer2020TAC-switched}, we formulate the problem of adapting dynamic locomotion to externally applied forces as a switching system among a collection of dynamic walking gait controllers regulated by a high-level supervisor; see Fig.~\ref{fig:block_diagram}. To decide which gait controller must be engaged, the supervisor incorporates a \ac{NN} that learns---with probabilistic generalization guarantees---how to interpret noisy information about the interaction forces as a directional indicator associated with an unknown intended trajectory. The low-level feedback control loop then executes the controller suggested by the supervisor.

The hierarchical structure of the proposed approach effectively \emph{decouples} locomotion stability from adaptation. The goal of stability is achieved in a tractable way by ensuring that the gait library contains controllers capable of generating stable gaits and that switching among them is sufficiently slow; see our previous work \cite{Veer2019ICRA,Veer2020TAC-switched} for details. On the other hand, the goal of adaptation to intended but unknown trajectories via physical interaction is realized by training the supervisor to recognize changes in the direction of the intended trajectory by examining the interaction force and the motion of the robot. 
%
%
In particular, we train the supervisor by minimizing the PAC-Bayes generalization bound, which results in a ``certificate of performance'' on novel leader trajectories. Note that adaptation is not part of the low-level control design, which focuses explicitly on realizing stable locomotion; instead, adaptation occurs at the high level, by selecting the controller which is more suitable based on  stride-to-stride (low-frequency) feedback to the supervisor. 
The rest of the paper particularizes the adaptive supervisory control scheme described above to the application of adapting dynamic locomotion to an unknown desired trajectory $\mathrm{p}_\mathrm{L}(t)$. Before we proceed, it is worth emphasizing here an important advantage of the proposed hierarchical framework:  \emph{modularity}. Although the supervisor's ability to adapt the system's behavior relies on properties of the individual controllers, it does \emph{not} rely on the specific low-level design details of how these properties are realized by the low-level controllers. An immediate benefit is that ``off-the-shelf'' feedback control design methods can be used to design the low-level gait controllers, as long as the resulting library can generate sufficiently rich locomotion behaviors to successfully accomplish the task at hand. 
%



\section{PAC-Bayes Switching Policies for Adaptation}
\label{sec:PAC-switching}

This section formalizes the main ideas that underlie the proposed framework; see Fig.~\ref{fig:block_diagram}.    

\vspace{-0.1in}
\subsection{From Limit-cycle Gaits to Dynamic Movement Primitives}

Dynamic bipedal walking can be modeled by distinguished periodic solutions---that is, \emph{limit cycles}---of hybrid robot models~\cite{westervelt2007feedback}. Using Poincar\'e's method~\cite{westervelt2007feedback}, the behavior of such systems, locally \emph{around} a limit cycle, can be naturally represented as a \ac{DMP}~\cite{Veer2019ICRA}.

In more detail, consider a collection of limit cycles $\{ \mathcal{O}_r ~|~ r \in \mathcal{R} \}$, where $\mathcal{R}$ is a finite index set. The limit cycles $\mathcal{O}_r$ are designed to capture walking gaits with different attributes; e.g., walking straight, turning with different angles, or other behaviors relevant to the task. Then, Poincar\'e's method~\cite{westervelt2007feedback} effectively associates each limit cycle $\mathcal{O}_r$ with a equilibrium (fixed) point of a discrete dynamical system that captures the stride-to-stride evolution of the robot during the corresponding walking gait. Let $\mathcal{X} \subset \mathbb{R}^n$ denote the state space of the robot and let $\mathcal{S} \subset \mathcal{X}$ be a surface transversal to the limit cycle $\mathcal{O}_r$ for all $r \in \mathcal{R}$; typically, $\mathcal{S}$ is selected to be the ground surface. Then, $\mathcal{O}_r$ can be represented by a fixed point $x^*_r \in \mathcal{S}$ of the discrete-time dynamical  system
\begin{equation} \label{eq:discrete-system}
    x_{k+1} = f^*_r(x_k)\enspace, ~~~~~~~ r \in \mathcal{R} 
\end{equation}
where $x \in \mathcal{S}$ denotes the state, $f^*_r : \mathcal{S} \to \mathcal{S}$ is the corresponding Poincar\'e map
and $x^*_r = f^*_r(x^*_r)$; see  \cite{westervelt2007feedback}.

With this construction, the behavior of the system locally around a limit-cycle walking gait $\mathcal{O}_r$ can be formalized as a 2-tuple containing the map $f^*_r$ and its fixed point $x^*_r$; i.e.
\begin{equation} \label{eq:dmps}
    \mathcal{G}_r = \{f^*_r, ~x^*_r \}\enspace, ~~~~~~~ r \in \mathcal{R} \enspace 
\end{equation}
%
which represents an \emph{attractor landscape}~\cite{ijspeert2013dynamical} that captures not only the nominal limit-cycle walking gait, but also the dynamics of the system around it. 
%
We consider each $\mathcal{G}_r$ as a \ac{DMP} and we refer to the collection $\mathbb{G} = \{\mathcal{G}_r ~|~ r \in \mathcal{R} \}$ as the library of the \ac{DMP}s available to the supervisor.
%

\subsection{The Supervisor: Adapting  via Switching}

Equipped with a library $\mathbb{G}$ of \ac{DMP}s, the supervisor processes incoming information about the robot and its environment, and decides which \ac{DMP}  must be implemented at the ensuing stride.
Here, the term ``environment'' encompasses effects that are external to the robot, yet influence the evolution of its state. For example, when the robot is tasked with tracking an unknown trajectory $\mathrm{p}_\mathrm{L}(t)$ representing the leading co-worker's intent, $\mathrm{p}_\mathrm{L}(t)$ will be considered as part of the environment. Other effects, such as random initial conditions, noisy measurements, model uncertainty, or workspace geometry can also be considered as parts of the environment. To emphasize the role of an environment $E$ on the robot's motion when primitive $\mathcal{G}_r$ is executed, we write 
\begin{equation} \label{eq:discrete-system-E}
    x_{k+1} = f_r(x_k; ~E)\enspace, ~~~~~~~ r \in \mathcal{R} 
\end{equation}
where $f_r : \mathcal{S} \times \mathcal{E} \to \mathcal{S}$ is the state update rule and $E$ is assumed to belong in some set $\mathcal{E}$ of environments. 
%
%
Comparing \eqref{eq:discrete-system-E} with \eqref{eq:discrete-system}, we can interpret \eqref{eq:discrete-system} as the evolution of the system in a \emph{nominal} environment $E^* \in \mathcal{E}$ for which the gait primitives are derived; i.e., $f^*_r(x) = f_r(x;~ E^*)$. However, different environments are encountered during task execution, causing the state to evolve according to \eqref{eq:discrete-system-E}. 

Now, the information available to the supervisor can be captured by a mapping $H: \mathcal{X} \times \mathcal{E} \to \mathcal{Y}$, that furnishes a partial observation, i.e., \emph{cue}, $y \in \mathcal{Y}$ from a state $x \in \mathcal{X}$ and an environment $E \in \mathcal{E}$. In the context of a biped following an unknown trajectory, such cues may include certain gait features---e.g., heading or speed---as well as measurements of the interaction forces developed during the task. Based on this information, the supervisor outputs a sequence $\sigma : \mathbb{Z_+} \to \mathcal{R}$ that maps the $k$-th stride of the robot to the index 
\begin{equation}\label{eq:switching}
r = \sigma(k)
\end{equation}
of the gait primitive $\mathcal{G}_r \in \mathbb{G}$ that is required at that stride. 
Equation ~\eqref{eq:switching} defines a switching signal 
and gives rise to the (perturbed) switched system with multiple equilibria~\cite{Veer2020TAC-switched},
\begin{equation}\label{eq:switching-environment}
    x_{k+1} = f_{\sigma(k)}(x_k; ~E)
\end{equation}
which describes the dynamics of the robot in response to the supervisor's sequence of decisions \eqref{eq:switching}. 

Before we elaborate on switching policies, a comment on safe operation is in order. Effectively, \eqref{eq:switching-environment} describes a system that ``sways'' among fixed points. Persistent switching causes \eqref{eq:switching-environment} to be in a ``perpetual'' transient phase, never converging to any of the underlying fixed points. Defining safety for systems like \eqref{eq:switching-environment}, and providing \emph{explicit} guarantees for safe operation, has been addressed in~\cite{Veer2019ICRA, Veer2020TAC-switched}, where we required that (i) each primitive in $\mathbb{G}$ is a (locally) exponentially stable fixed point, and (ii) switching among primitives is sufficiently slow. Under these conditions, possible divergent behavior due to switching is suppressed by the exponential convergence between switches. It was proved in~\cite{Veer2020TAC-switched} that the state of \eqref{eq:switching-environment} is trapped within a compact region of the state space, the size of which can be adjusted to ensure practically stable operation. We will not delve deeper into this issue here, since our goal in this paper is to devise adaptive switching policies for the supervisor; we just mention that theoretical tools are available to ensure that these policies are provably safe; see~\cite{Veer2019ICRA, Veer2020TAC-switched}.  

\subsection{Learning Provably Generalizable Switching Policies}

In the context of adapting to intention via interaction, we are concerned with learning policies $\pi : \mathcal{Y} \to \mathcal{R}$ in a policy space $\Pi$ that map ``cues'' $y \in \mathcal{Y}$ regarding the state of the system and its environment to the index $r \in \mathcal{R}$ of the \ac{DMP} $\mathcal{G}_r$ in $\mathbb{G}$ that is ``best'' to employ. Importantly, our goal in this work is to learn switching policies, which, given a dataset of environment instances, generalize with \emph{provable guarantees} to novel environments. To achieve this, we will utilize \ac{PAC} Bayes theory, which is known to provide strong generalization bounds in supervised learning~\cite{mcallester1999some,dziugaite2017computing}. 

We begin by assuming the availability of a cost function $C: \Pi \times \mathcal{E} \to [0,1]$, which captures critical aspects of the task and can be used to assess the ``quality'' of employing a policy $\pi \in \Pi$ in a particular environment $E \in \mathcal{E}$. Note that there is no loss of generality in constraining the co-domain of the cost function to $[0,1]$; indeed, any bounded cost function could be used as long as it is scaled with a suitable constant. 

Next, we assume that there is a distribution $\mathcal{D}$ over the space  $\mathcal{E}$ of possible environments; this distribution reflects the underlying stochastic mechanism by which an environment is encountered by the system. It is important to emphasize that we do \emph{not} assume knowledge of $\mathcal{D}$.  
In this setting, our objective is to learn policies that minimize the expected cost across environments generated according to $\mathcal{D}$,
\begin{equation} \label{eq:problem-definition}
    \min_{\pi \in \Pi} C_\mathcal{D}(\pi) = \min_{\pi \in \Pi} \mathop{\mathbb{E}}_{E \thicksim \mathcal{D}}[C(\pi;~E)] \enspace.
\end{equation}

To formulate the optimization problem \eqref{eq:problem-definition} so that \ac{PAC}-Bayes theory can be applied, we will randomize the policy space $\Pi$. This is done by assuming a distribution $P$ over $\Pi$ according to which individual policies can be selected. In this setting, when the robot encounters an environment $E$, the supervisor randomly selects a policy from $P$ and applies it to decide which gait primitive should be engaged in the forthcoming stride. 
With this modification, if $\mathcal{P}$ denotes the space of distributions defined over $\Pi$, our goal becomes to learn policy distributions $P \in \mathcal{P}$ that realize the minimum
\begin{equation}\label{eq:cost-opt}
     C^{\star} = \min_{P \in \mathcal{P}} C_{\mathcal{D}}(P) = \min_{P \in \mathcal{P}} \displaystyle \mathop{\mathbb{E}}_{E \thicksim \mathcal{D}} \left[ \displaystyle \mathop{\mathbb{E}}_{\pi \thicksim P} [C({\pi};E)] \right] \enspace.
 \end{equation}

However, as was mentioned above, the distribution $\mathcal{D}$ is not known, and thus the expectation over $\mathcal{D}$ in \eqref{eq:cost-opt} cannot be explicitly computed. Yet, \emph{indirect} knowledge about $\mathcal{D}$ can be obtained by sampling the space of environments $\mathcal{E}$, resulting in datasets corresponding to (finite) collections of environments $D = \{E_1,E_2,...,E_N\}$. Then, the expectation over $\mathcal{D}$ can be approximated by the empirical average
\begin{equation}\label{eq:cost-empirical}
    C_D(P) = \frac{1}{N} \displaystyle \mathop{\sum}_{E \in D} \displaystyle \mathop{\mathbb{E}}_{\pi \thicksim P} [C(\pi; E)] \enspace.
\end{equation}
The \ac{PAC}-Bayes generalization framework \cite{mcallester1999some, rivasplata2019pac} provides a computable upper bound on the expected true cost $C_\mathcal{D}(P)$ involved in \eqref{eq:cost-opt} in terms of the empirical cost $C_D(P)$ in \eqref{eq:cost-empirical}. 

To apply the \ac{PAC}-Bayes framework, we assume the availability of a ``prior'' distribution $P_0 \in \mathcal{P}$ \emph{before} observing any data. Note that $P_0$ is \emph{not} a Bayesian prior; that is, the correctness of the \ac{PAC}-Bayes bound is not subject to the correctness of the prior, thus providing enhanced flexibility in the choice of $P_0$. The crucial benefit of this flexibility is that (partial) knowledge of the problem structure can be embedded as inductive bias in the learning framework without compromising the correctness of the bounds. 
%
Theorem~\ref{thm:PAC-Bayes} provides an explicit expression for the \ac{PAC}-Bayes upper bound on the true cost $C_\mathcal{D}(P)$, which does \emph{not} rely on the explicit knowledge of $\mathcal{D}$ and can therefore be computed using data samples; for a proof see~\cite{veer2020probably}. 
\begin{theorem}[adapted from \cite{veer2020probably}, \cite{rivasplata2019pac}]\label{thm:PAC-Bayes}
Let $\delta \in (0,1)$, $D \sim \mathcal{D}^N$ be a multisample $D = \{E_1,E_2,...,E_N\}$ of $N$ training environments drawn in an \ac{iid} fashion from $\mathcal{E}$ according to $\mathcal{D}$, and $P_0\in\mathcal{P}$ be a prior distribution on the space of policies $\Pi$. Then, with probability at least $1 - \delta$, for any posterior distribution $P\in\mathcal{P}$, the following inequality holds:
\begin{multline}
    C_\mathcal{D}(P) \le C_\mathrm{QPAC}(P,P_0) := \\ \left(\sqrt{C_D(P) +R(P,P_0) } + \sqrt{R(P,P_0)}\right)^2 \label{bound-QPAC}
\end{multline}
in which $C_D(P)$ is given by \eqref{eq:cost-empirical} and $R(P,P_0)$ is defined as
\begin{equation}\label{eq:regularizer}
    R(P,P_0)=\frac{\mathrm{KL}(P||P_0)+\log(\frac{2\sqrt{N}}{\delta})}{2N}
\end{equation}
where $\mathrm{KL}(P||P_0)$ denotes the Kullback-Leibler divergence (relative entropy) from $P_0$ to $P$. 
\end{theorem}

The importance of Theorem~\ref{thm:PAC-Bayes} is that we can find a posterior policy distribution $P$ by minimizing the bound $C_\mathrm{QPAC}$, which consists of two terms: (i) the empirical cost $C_D(P)$, and (ii) the regularizer $R(P,P_0)$. Intuitively, minimizing $C_D(P)$ tries to ``fit" the posterior $P$ to the training data $D$, while the regularizer $R$ penalizes over-fitting.
Then, equipped with a prior $P_0$ and samples $D$ of environments from $\mathcal{E}$, finding a posterior distribution that minimizes the \ac{PAC}-Bayes bound \eqref{bound-QPAC} can be done in a computationally tractable manner using convex optimization tools; see Section~\ref{subsec:training}.

\begin{remark}\label{rem:prior-ind}
Any approach can be employed to obtain a suitable prior distribution $P_0$; the only criterion that must be satisfied is the independence of $P_0$ from the training dataset $D$ that is used for the application of Theorem~\ref{thm:PAC-Bayes}. The flexibility in choosing $P_0$ allows us to design informative priors by leveraging highly parallelizable \ac{ES}, as described in Section~\ref{subsec:training} below. 
\end{remark}

\section{Task Modeling and Gait Library}
\label{sec:task-model}

This section applies the framework described above to the problem of a dynamic biped  following an unknown desired trajectory through physical interaction with a leader.

\subsection{Task Model: Intention via Interaction}

It is assumed that the leader's intention over a time interval $[0, T]$ with $T > 0$ is encoded in a sufficiently smooth desired trajectory $\mathrm{p}_\mathrm{L}(t)$ where $t \in [0, T]$. The biped does not explicitly know $\mathrm{p}_\mathrm{L}(t)$; instead, it perceives the intended trajectory via a force $F_\mathrm{e}(t)$ developed at the port of interaction between the leader and the biped. As is common in human-robot physical interaction, an impedance model is implemented to translate $\mathrm{p}_\mathrm{L}(t)$ to the force experienced by the biped; i.e.,  
\begin{equation}\label{eq:impedance}
    F_\mathrm{e}(t) = K_{\mathrm{L}}\Big( \mathrm{p}_\mathrm{L}(t) - \mathrm{p}_\mathrm{R}(t) \Big) + N_{\mathrm{L}}\Big(\dot{\mathrm{p}}_\mathrm{L}(t) - \dot{\mathrm{p}}_\mathrm{R}(t) \Big)
\end{equation}
where $K_{\mathrm{L}}$ and $N_{\mathrm{L}}$ are the stiffness and damping matrices, respectively, and $\mathrm{p}_\mathrm{R}$ denotes the point on the robot at which the force is applied. 
In the case where the robot interacts with a human via its arms, $\mathrm{p}_\mathrm{R}$ corresponds to the position of the end effector; see~\cite{motahar2017steering} for details. To simplify the exposition without changing the essential features of the problem, we will assume that the port of interaction is a point on the robot's torso as shown in Fig.~\ref{fig:biped_model}. 

\subsection{Robot Model and Controller Design}

\begin{figure}[t!]
\begin{centering}
\includegraphics[width=1\columnwidth]{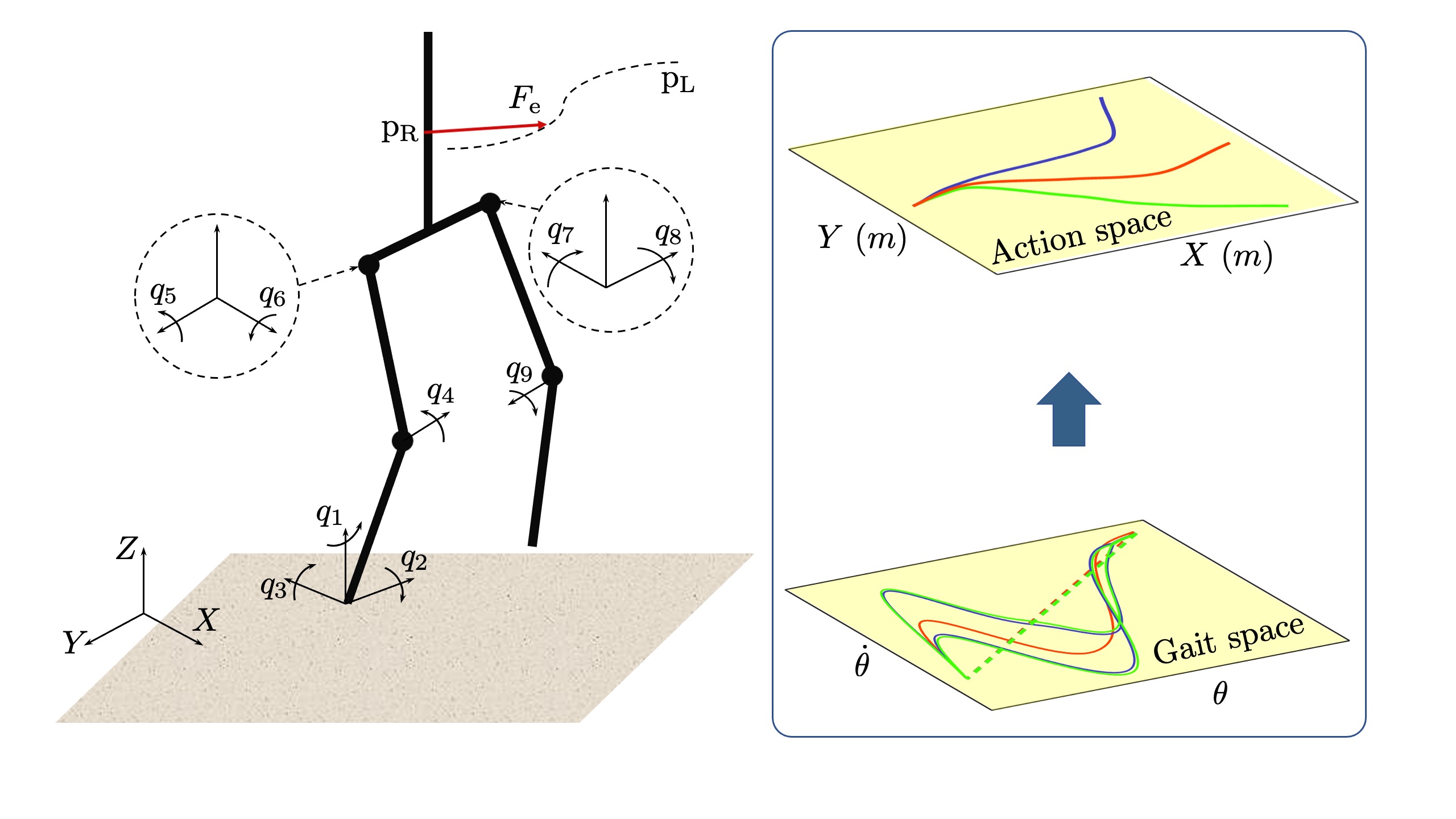}
\par\end{centering}
\vspace{-0.1in}
\caption{(Left) A 3D bipedal robot model with a choice of coordinates. The biped experiences an interaction force $F_\mathrm{e}(t)$ that carries information about a desired, yet unknown, trajectory $\mathrm{p}_\mathrm{L}$. (Right) On the low level, a gait controller design method is employed to extract a library of limit-cycle gait primitives. On the high level, each gait primitive generates an action that represents suitable displacements.}
\vspace{-0.2in}
\label{fig:biped_model} 
\end{figure}

One of our objectives is to show that our approach can harness \emph{existing} locomotion control design methods without the need of major modifications. Thus, we adopt the 3D bipedal robot model of Fig.~\ref{fig:biped_model}, for which effective walking controllers are available in the relevant literature~\cite{Shih2012, motahar2017steering}, noting that other models or controllers can also be used. 

In our setting, bipedal walking is characterized by a sequence of alternating left and right leg support phases punctuated by double support phases. In single support, the model possesses nine \ac{DOF}, which can be described by the coordinates $q = (q_1,q_2,...,q_9)$; see Fig.~\ref{fig:biped_model}. We assume that all \ac{DOF}s except from the yaw $q_1$ and pitch $q_2$ angles of the foot are actuated. As in~\cite{Shih2012}, double support phases are assumed to be instantaneous and are modeled as impact events as in~\cite[Chapter 3]{westervelt2007feedback}.

Due to the non-trivial length of the hip, the equations of motion for the left and right (single) support phases differ, but both can be written as systems with impulse effects,
\begin{eqnarray} \label{eq:hybrid}
  \mathcal{H}:
  \begin{cases}
    \begin{aligned}
        \dot{x} &= \alpha(x)+\beta(x)u+\beta_{{\rm e}}(x)F_{{\rm e}}, & x\notin\mathcal{S} \\
        x^+ &= \Delta(x^{-}), & x^{-}\in\mathcal{S}
    \end{aligned}
  \end{cases}
\end{eqnarray}
where $x = [q^\mathsf{T},\dot{q}^\mathsf{T}]^\mathsf{T}$ is the state and the vector fields $(\alpha, \beta, \beta_\mathrm{e})$ describe the single support dynamics under the inputs $u \in \mathbb{R}^7$ and the interaction force $F_\mathrm{e}$ modeled according to \eqref{eq:impedance}. In \eqref{eq:hybrid}, $\mathcal{S}$ is the ground surface and $\Delta$ maps the state $x^-$ prior to impact to the state $x^+$ right after impact. 

Next, we turn our attention to the design of a low-level controller for walking. Here, we use the \ac{HZD} framework as in~\cite{westervelt2007feedback}, assuming that measurements $\tilde{F}_\mathrm{e}$ of the interaction force $F_\mathrm{e}$ are available. The end result is a feedback law of the form $u = \Gamma (x, \tilde{F}_\mathrm{e})$
for each of the left and right leg support phases \eqref{eq:hybrid}. Due to space constraints, we omit the details of controller design, which can be found in~\cite{Shih2012, motahar2017steering}. We only mention that \eqref{eq:hybrid} in closed loop with $\Gamma$ results in a \emph{forced} system with impulse effects~\cite{veer2019TAC-ISS} representing the low-level dynamics of the biped. 
%

\subsection{Gait primitives and actions}

\begin{figure*}[ht]
\begin{centering}
\includegraphics[width=0.99\textwidth]{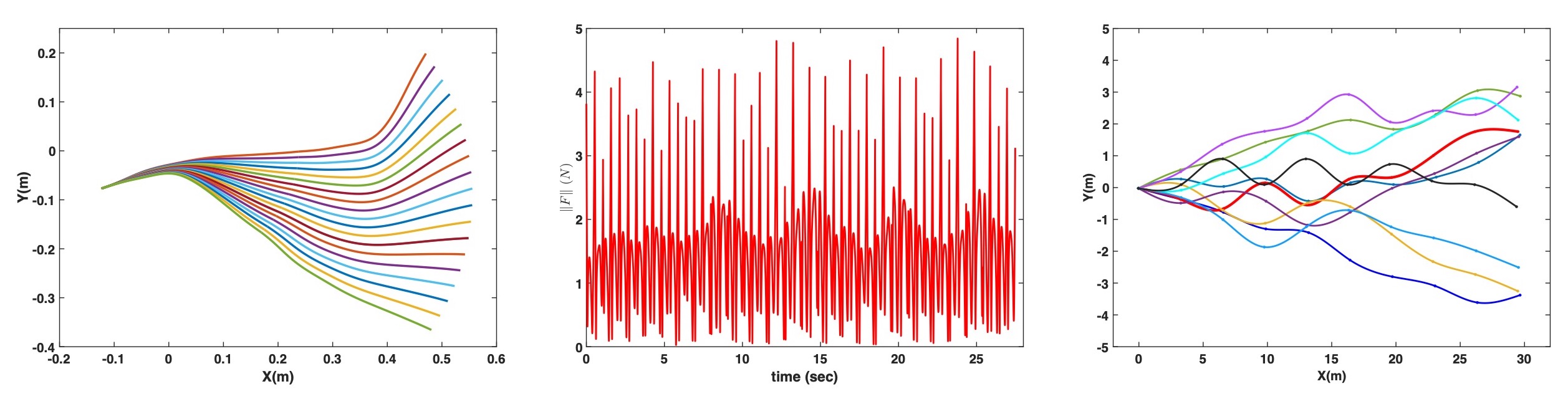}
\par\end{centering}
\caption{(Left) Displacements (actions) in Cartesian space corresponding to the gaits in $\mathbb{G}$; no external force is considered here. (Middle) Magnitude of the noisy force $\tilde{F}_\mathrm{e}(t)$ corresponding to the bold red trajectory in the right figure. (Right) Sampled trajectories $\mathrm{p}_\mathrm{L}(t)$.}
\vspace{-0.2in}
\label{fig:gait_primitives} 
\end{figure*}

To extract a collection of limit-cycle walking gaits $\mathbb{G} = \{ \mathcal{G}_r ~|~ r \in \mathcal{R} \}$, we design a finite family of feedback controllers $u = \Gamma_r(x,\tilde{F}_\mathrm{e})$ indexed by $r \in \mathcal{R}$. 
%
In the absence of the external force, these controllers generate  (locally exponentially) stable limit cycles corresponding to  walking gaits with stride-by-stride turning angles in the range $[-45^\circ, 45^\circ]$ with $5^\circ$ increments; see Fig. \ref{fig:gait_primitives} (left). These gait primitives are obtained by solving a nonlinear optimization problem as in~\cite{westervelt2007feedback}. Due to space limitations, we skip the details associated with the control design; we only mention that the closed-loop dynamics of the system evolving under $\Gamma_r$ in the absence of any interaction defines the gait primitives $\mathcal{G}_r = \{f^*_r, x^*_r \}$ defined in  Section~\ref{sec:PAC-switching}.

\section{Learning to Adapt via Physical Interaction}
\label{sec:results}

This section provides details on training switching policies that provably generalize to novel environments.

\vspace{-0.1in}
\subsection{Environment Generation}

Over the interval $[0,~ T]$, the evolution of the closed-loop system depends on the initial condition $x_0 = x(0)$ and on the unknown desired trajectory $\{\mathrm{p}_\mathrm{L}(t) ~|~ t \in [0, T]\}$, which are considered as part of the environment of the system. Furthermore, as was mentioned above, we assume that only noisy measurements $\{\tilde{F}_\mathrm{e}(t) ~|~ t \in [0, T]\}$ of the interaction force \eqref{eq:impedance} are available; see Fig. \ref{fig:gait_primitives} (middle) for an example of a typical forcing pattern. Noise will also be considered as part of the environment, an instance of which is defined by
\begin{equation}\label{eq:environment-set}
    E = \big\{ \{x_0, \tilde{F}_\mathrm{e}(t),  \mathrm{p}_\mathrm{L}(t)\} ~|~ t \in [0, T] \big\} 
\end{equation}
which belongs in the space $\mathcal{E}$ of 3-tuples composed by the initial conditions and the \emph{functions} describing the desired trajectory and the measurements of the interaction forces.

Next, we describe the mechanism that defines the distribution $\mathcal{D}$ by which random environment instances \eqref{eq:environment-set} are sampled from the space $\mathcal{E}$ of possible environments. 
In more detail, the initial conditions in $E$ are selected to be those of the walking straight fixed point with randomly selected yaw angle $\tilde{q}_1 = q_1 + w_q$ with $w_q \thicksim \mathcal{N}(0, \sigma_q)$. The measurements $\{ \tilde{F}_\mathrm{e}(t) ~|~ t \in [0, T] \}$ of the force that are available to the low-level controllers are assumed to be corrupted by white noise; i.e., $\tilde{F}_\mathrm{e}(t) = F_\mathrm{e}(t) + w_F$, where $w_F \thicksim \mathcal{N}(0, \sigma_F)$ and $F_\mathrm{e}(t)$ is given by \eqref{eq:impedance}.
Finally, the desired trajectories $\{ \mathrm{p}_\mathrm{L}(t) ~|~ t \in [0, T] \}$ are generated by smoothening sequences of line segments of equal length with randomly selected slopes from a uniform distribution over the set  $[-15^\circ, 15^\circ]$. 
%

Our choice to represent the desired trajectories $\mathrm{p}_\mathrm{L}(t)$ as above is motivated by collaborative object transportation tasks that involve pairs of co-workers in which one assumes the role of the leader~\cite{motahar2017steering, Veer2019ICRA}. In such tasks, the leader is assumed to know where the object must be transported, and typically plans a smooth  trajectory towards the goal location, occasionally changing directions to avoid possible obstacles. By modeling $\mathrm{p}_\mathrm{L}(t)$ this way, during training the supervisor effectively learns how to interpret noisy measurements of interaction forces in terms of changing directions in the intended trajectory. Finally, note that a wide variety of functions $\mathrm{p}_\mathrm{L}(t)$ can be generated this way; Fig. \ref{fig:gait_primitives} (right) shows sampled desired trajectories.


\subsection{Policy Parameterization}

To find a \ac{PAC}-Bayes policy, we parameterize the space of policies using a \ac{NN} architecture with weights $w \in \mathbb{R}^d$. The policy is a \ac{NN} with $d=689$ parameters which consists of an input layer, two hidden layers and an output layer with $n_i$, 10, 20, and $n_o$ neurons in each layer, respectively. The hidden layers are activated using an exponential linear unit (\textsc{elu}) activation function
while the output layer is activated with the Softmax activation function, which assigns the gait primitive scores as outputs.

The \ac{NN} receives at its input a set of partial observations $y \in \mathcal{Y}$ that capture relevant gait and interaction features, denoted by $\zeta$ and $(\Phi^\mathrm{x}_\mathrm{e}, \Phi^\mathrm{y}_\mathrm{e})$, respectively. These features are functions of the robot's state $x$ and its environment $E$; i.e., 
\begin{equation}\nonumber
    y = H(x; E) = 
    \begin{bmatrix}
    \zeta(x; E) & \Phi^\mathrm{x}_\mathrm{e}(x; E) & \Phi^\mathrm{y}_\mathrm{e}(x; E)
    \end{bmatrix}
\end{equation}
More specifically, let $[t_{k-1}, t_k]$ be the duration of the $k$-th stride. Then, the gait features capture aspects of the geometry of the walking motion at the end of the $k$-th stride i.e. at time $t_k$. They include the heading angle $q_1$ and the angle $\theta(q) = -q_2 - q_4/2$ of the line connecting the hip and the foot of the support leg, as well as their rates; that is\footnote{Intuitively, $(q_1, \dot{q}_1)$ provide information about the robot's heading and $(\theta,\dot{\theta})$ about its stride length and frequency.}, $\zeta = (q_1, \theta, \dot{q}_1, \dot{\theta})$. 
The interaction features capture the effect of the external force over the duration $[t_{k-1}, t_k]$ of the $k$-th stride, and are
\begin{equation}\nonumber
   \Phi^\mathrm{x}_\mathrm{e}(x; E) = \int^{t_k}_{t_{k-1}} \tilde{F}^\mathrm{x}_\mathrm{e}(t) dt~~~\text{and}~~~ \Phi^\mathrm{y}_\mathrm{e}(x; E) = \int^{t_k}_{t_{k-1}} \tilde{F}^\mathrm{y}_\mathrm{e}(t) dt
\end{equation}
where $\tilde{F}_\mathrm{e}(t)$ are the noisy measurements of the interaction force as explained above. The \ac{NN} processes this information and assigns a score to each gait primitive in $\mathbb{G}$, based on which the supervisor selects a suitable walking gait.
%


\subsection{Training}
\label{subsec:training}

The training pipeline consists of two stages. 
In the first stage, an inductive bias is extracted in the form of a probability distribution $\hat{P}$ over the space of policies. Loosely speaking, $\hat{P}$ reflects the ``quality'' of the policies generated by the \ac{NN}. Choosing such distribution for \ac{NN}s is often not intuitive; to remedy this problem, we adopt the approach in~\cite{veer2020probably} and employ \ac{ES} to compute $\hat{P}$ using a dataset $\hat{D} \sim \mathcal{D}^{\hat{N}}$ of $\hat{N}$ training environments.
In the second stage, we leverage $\hat{P}$ to extract an informative prior distribution, which is then used to optimize the \ac{PAC}-Bayes bound of Theorem~\ref{thm:PAC-Bayes}. To do this, a dataset $D \sim \mathcal{D}^N$ of $N$ training environments is used; the dataset $D$ is sampled independently from $\hat{D}$.

\subsubsection{Extracting inductive bias on the policy space}
\label{subsubsec:P_hat}
Our objective here is to uncover a useful inductive bias on the performance of the policies generated by the \ac{NN}. Specifically, a probability distribution $\hat{P}$ will be obtained that is ``peaked'' around policies that, on average, perform well on environments drawn from $\mathcal{D}$. To evaluate the performance of a policy $\pi$ applied in an environment $E$, we use the cost
\begin{equation}\label{eq:cost-prior}
    \hat{C}(\pi;E) = \frac{1}{L} \int_0^T (e_\mathrm{p}^2(t) +  e_{\phi}^2(t)) ~dt
\end{equation}
in which $L$ is the distance traveled by the leader over the interval $[0,T]$ and $e_\mathrm{p}$ and $e_{\phi}$ correspond to position and orientation errors~\cite[Chapter~5]{LaValle2006book}. In more detail,
\begin{align}
    e_\mathrm{p} &= \|\mathrm{p}_\mathrm{L} - \mathrm{p}_\mathrm{R}\| \nonumber \\
    e_{\phi} &= \sqrt{(\cos{\phi_\mathrm{L}} - \cos{\phi_\mathrm{R}})^2 + (\sin{\phi_\mathrm{L}} - \sin{\phi_\mathrm{R}})^2} \nonumber~\enspace
\end{align}
where dependence on $t$ has been omitted, and $\phi_\mathrm{L}$ and $\phi_\mathrm{R}$ are the slopes of $\mathrm{p}_\mathrm{L}$ and $\mathrm{p}_\mathrm{R}$; see Fig.~\ref{fig:biped_model}. Essentially, \eqref{eq:cost-prior} assesses the ``quality'' of a policy based on the error of the biped's position and orientation from the intended trajectory.

To obtain the distribution $\hat{P}$, we will restrict our attention to the family of multivariate Gaussian distributions with diagonal covariance matrices; let $\mu \in \mathbb{R}^d$ be the mean and $\sigma \in \mathbb{R}^d$ the square root of the diagonal elements of the covariance matrix $\Sigma \in \mathbb{R}^{d \times d}$. Then, we sample $\hat{N}$ environments $\hat{D}$ from $\mathcal{E}$ and minimize the resulting empirical cost $\hat{C}_{\hat{D}}(\hat{P})$ with respect to the parameters $\psi:=(\mu,\sigma) \in \mathbb{R}^{2d}$. 
%
The process requires the computation of the gradient
\begin{equation} \nonumber
    \nabla_{\psi}\hat{C}_{\hat{D}}(\hat{P}) = \frac{1}{\hat{N}}\displaystyle \mathop{\sum}_{E \in \hat{D}} \nabla_{\psi} \mathop{\mathbb{E}}_{w \thicksim \hat{P}} [\hat{C}(\pi_w; E)]
\end{equation}
which, following~\cite{wierstra2014natural, veer2020probably} can be decomposed as
\begin{align}
\nonumber
    \nabla_{\mu} \!\mathop{\mathbb{E}}_{w \thicksim \hat{P}}[\hat{C}(\pi_w; E)] &= \!\mathop{\mathbb{E}}_{\epsilon \thicksim \mathcal{N}(0, I)} [\hat{C}(\mu \!+\! \sigma \!\odot\! \epsilon; E) \epsilon] \!\oslash\! \sigma 
    \\
    \nonumber
    \nabla_{\sigma} \!\mathop{\mathbb{E}}_{w \thicksim \hat{P}}[\hat{C}(\pi_w; E)] &= \!\mathop{\mathbb{E}}_{\epsilon \thicksim \mathcal{N}(0, I)} [\hat{C}(\mu \!+\! \sigma \!\odot\! \epsilon; E) (\epsilon \odot \epsilon \!-\! \mathbf{1})] \!\oslash\! \sigma
    \end{align}
where $\odot$ and $\oslash$ denote the Hadamard (elementwise) product and division, respectively, and $\mathbf{1}$ is the $d$-dimensional vector with $1$s. These expressions are used to estimate the gradient using Monte Carlo simulations over $2\hat{m}$ policies sampled from $\mathcal{N}(\mu, \Sigma)$.
More details regarding the implementation of \ac{ES} can be found in \cite{wierstra2014natural, veer2020probably}. We only mention here that we use antithetic sampling to reduce variance in the gradient estimates; that is, we always sample policies in $\epsilon$ and $-\epsilon$ pairs as detailed in \cite{veer2020probably}. 
With these estimates, the parameters $\psi$ are updated in a gradient descent fashion according to $\psi_{t+1}\leftarrow \psi_t - \eta\nabla_{\psi}C_{\hat{D}}(\hat{P})$, where $\eta$ is the learning rate. 
The outcome of the training process for the distribution $\hat{P}$ is the values $\hat{\psi} = (\hat{\mu},\hat{\sigma})$, which will be used to extract an informative prior for optimizing the \ac{PAC}-Bayes bounds \eqref{bound-QPAC}. 

\subsubsection{Computing the PAC-Bayes policy} 
\label{subsubsec:PAC}
To provide an intuitive interpretation of the \ac{PAC}-Bayes bound \eqref{bound-QPAC}, we define a cost function that penalizes policies based on the fraction of the interval $[0,T]$ over which the biped violates a tube of radius $r$ around the intended trajectory $\mathrm{p}_\mathrm{L}(t)$. In other words, the more a switching policy causes the biped to wander outside a pre-specified tube around the desired trajectory, the larger the cost associated with that policy is. Mathematically,
\begin{equation}\label{eq:cost-PAC}
    C(\pi,E) = \frac{1}{T} \int_0^T \mathbbm{1}_{\{\|e_\mathrm{p}\|\geq r\}} ~dt
\end{equation}
where $\|e_\mathrm{p}\| = \|\mathrm{p}_\mathrm{L} - \mathrm{p}_\mathrm{R}\|$, and $\mathbbm{1}_{A}$ denotes the indicator function for a given subset $A$. By definition, $C \in [0,1]$.

Next, a suitable prior $P_0$ must be selected for the purpose of optimizing \eqref{bound-QPAC}. One possible choice for $P_0$ is the distribution $\hat{P}$, which favors policies that perform better according to \eqref{eq:cost-prior}. Indeed, by Remark~\ref{rem:prior-ind}, using different cost functions for extracting an informative prior and for establishing the \ac{PAC}-Bayes generalization guarantees is possible.
However, choosing $\hat{P}$ as $P_0$ and minimizing the bound \eqref{bound-QPAC} with respect to the distribution $P$ results in a high-dimensional optimization problem over the policy space $\mathbb{R}^d$; here $d=689$. 

To overcome this issue, we discretize the policy space $\mathbb{R}^d$ by sampling $m$ policies from $\mathbb{R}^d$ according to $\hat{P}$ in an \ac{iid} fashion. This way, inductive bias is embedded in the resulting \emph{finite} policy space $\Pi:= \{\pi_1,\pi_2,...,\pi_m\}$, which will be used to optimize the \ac{PAC}-Bayes bounds \eqref{bound-QPAC}. In this setting, we choose as the prior distribution\footnote{Notation: Continuous probability distributions are denoted by uppercase letters while their discrete counterparts are denoted by lowercase letters.} $p_0$ over $\Pi$ to be the (discrete) uniform distribution. 
Working with discrete distributions has the benefit that the $\mathrm{KL}$ divergence is a convex function, allowing us to express the resulting optimization as a convex program~\cite{veer2020probably}. Furthermore, discrete distributions result in a significantly lower complexity for the space of policy distributions, thus yielding tighter bounds.

Now, to optimize the \ac{PAC}-Bayes bound \eqref{bound-QPAC}, we sample $N$ environments $D = \{E_i\}_{i=1}^N$ from $\mathcal{E}$ according to $\mathcal{D}$. For each $E_i$, we evaluate the cost \label{eq:cost-PAC-Bayes} associated with each policy $\pi_j$ in $\Pi$ and form the cost matrix $C_{ij} = C(\pi_j;E_i)$, with $1 \leq i \leq N$ and $1 \leq j \leq m$. Then, the average cost $C_j = \frac{1}{N}\sum^N_{i=1} C_{ij}$ of deploying policy $\pi_j$ across all  environments in $D$ can be stacked to form a cost vector $\bar{C} \in \mathbb{R}^m$. Then, $C_D(p)$ in \eqref{bound-QPAC} can be expressed linearly in $p$ as $\bar{C} p$, and following~\cite{veer2020probably}, minimizing the upper bound $C_\mathrm{QPAC}(p)$ can be written as
\begin{align}\label{eq:opt-REP}
    &\displaystyle \mathop{\min}_{p \in \mathbb{R}^m} \left(\sqrt{\bar{C}p +R(p,p_0) } + \sqrt{R(p,p_0)}\right)^2 \\ 
    &\mathrm{subject ~to}~~~\displaystyle \mathop{\sum}_{i=1}^{m} p_i = 1,~0 \le p_i \le 1 \label{eq:opt-REP-2}
\end{align}
where $R(p,p_0)$ is computed by \eqref{eq:regularizer} using $p$ and $p_0$ and closed-form expressions for $\mathrm{KL}$ divergence. 
Following \cite{majumdar2020pac},~\eqref{eq:opt-REP}-\eqref{eq:opt-REP-2} can be converted to a relative entropy program, an efficiently solvable class of convex programs. 

\vspace{-0.05in}
\subsection{Results and Interpretation}

We consider the scenario in which a dynamic biped  adapts its walking pattern based on physical interaction so that it follows an unknown intended trajectory $\mathrm{p}_\mathrm{L}(t)$ over the interval $[0, T]$. The \ac{NN} in the supervisor effectively learns switching policies among dynamic gait primitives that provably generalize well when the biped is presented with environments not encountered during the training phase.  

In the first stage of the proposed training pipeline, the distribution $\hat{P}$ is obtained using \ac{ES} as explained in Section~\ref{subsubsec:P_hat}; the relevant hyperparameters are given in Table~\ref{table:hparam}. The $\psi$ parameter updates are performed on minibatches of size 20 out of the $\hat{N}=500$ environments that are used.
Then, $\hat{P}$ is used to embed inductive bias in reducing the policy space to the finite collection $\Pi$ of $m=20$ policies. To obtain the \ac{PAC}-Bayes bound, we introduce a tube of radius $r=0.5 \mathrm{m}$ around $\mathrm{p}_\mathrm{L}(t)$ for computing~\eqref{eq:cost-PAC} and apply Theorem~\ref{thm:PAC-Bayes} in a discrete probability setting, as explained in Section~\ref{subsubsec:PAC}. We select $p_0$ to be the uniform probability distribution over the reduced (finite) policy space $\Pi$ and choose  $\delta = 0.01$. The optimized \ac{PAC}-Bayes bound for different numbers of training environments are given in Table \ref{table:pac-bayes}.
Obtaining\footnote{Training was performed on a desktop with 3.5 GHz Xeon W-2265 CPU, 12 cores, 64 GB RAM, and a 16 GB NVIDIA Quadro RTX 5000 GPU.} a meaningful $\hat{P}$ is the most challenging task in terms of computational time. Training the \ac{PAC}-Bayes policy takes $\thicksim$70 hours to compute the cost matrix on 1000 environments with 20 policy samples and $\thicksim$1 sec to solve \eqref{eq:opt-REP}-\eqref{eq:opt-REP-2}.

\vspace{-0.1in}
\begin{table}[h!]
\renewcommand{\arraystretch}{1.3}
\caption{\small Hyperparameters used in our two-stage training pipeline}
\centering
\begin{tabular}{|c | c | c |c c||c|}
\hline
\multicolumn{5}{|c||}{\bfseries Inductive bias} & \bfseries \ac{PAC}-Bayes \\
\hline\hline
Initial&\#Envs & \# $\epsilon$ & \multicolumn{2}{|c||}{learning rate $(\eta)$}& \# Policy \\
Dist.& $\hat{N}$ & $\hat{m}$ & $\mu$ &$\text{log}(\sigma \odot \sigma)$& $m$ \\
\hline
$\mathcal{N}(0,I)$ & 500 & 2 & 0.1 & 0.01 & 20\\
\hline
\end{tabular}
\label{table:hparam}
\end{table}

\begin{figure*}
     \centering
     \begin{subfigure}{\columnwidth}
         \centering
         \includegraphics[trim=135 0 165 0, clip, width=0.99\columnwidth]{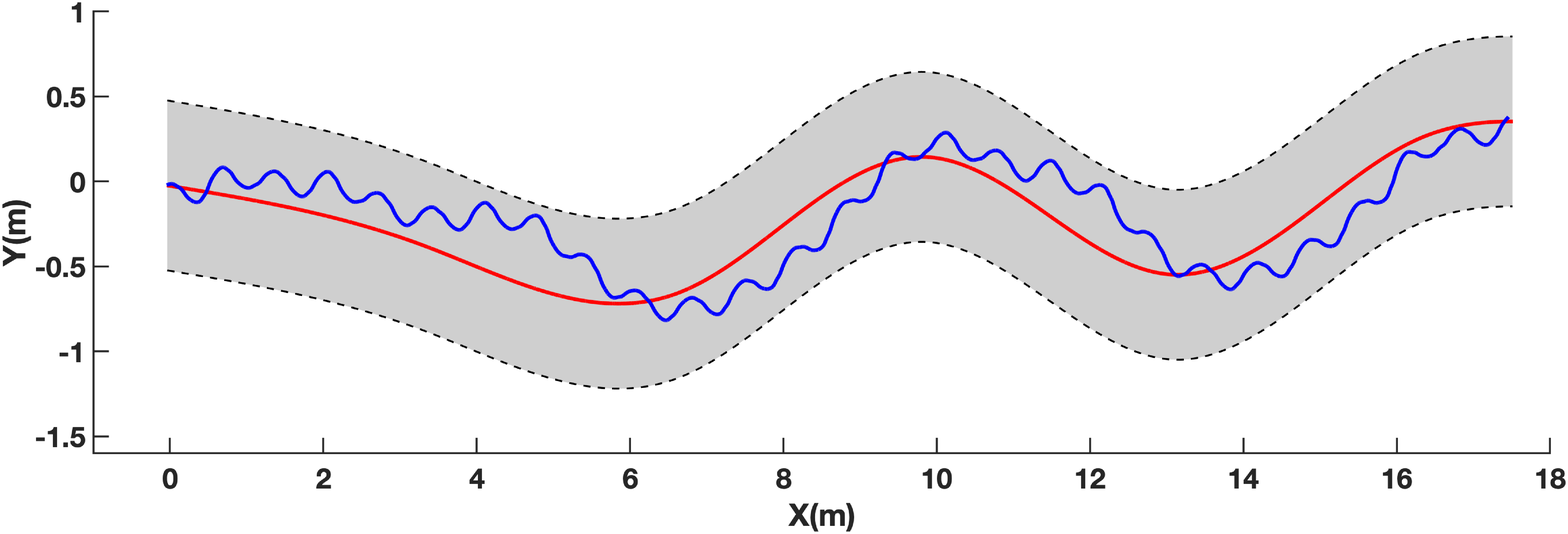}
     \end{subfigure}
     \begin{subfigure}{\columnwidth}
         \centering
         \includegraphics[trim=75 0 80 0,clip,width=0.99\columnwidth]{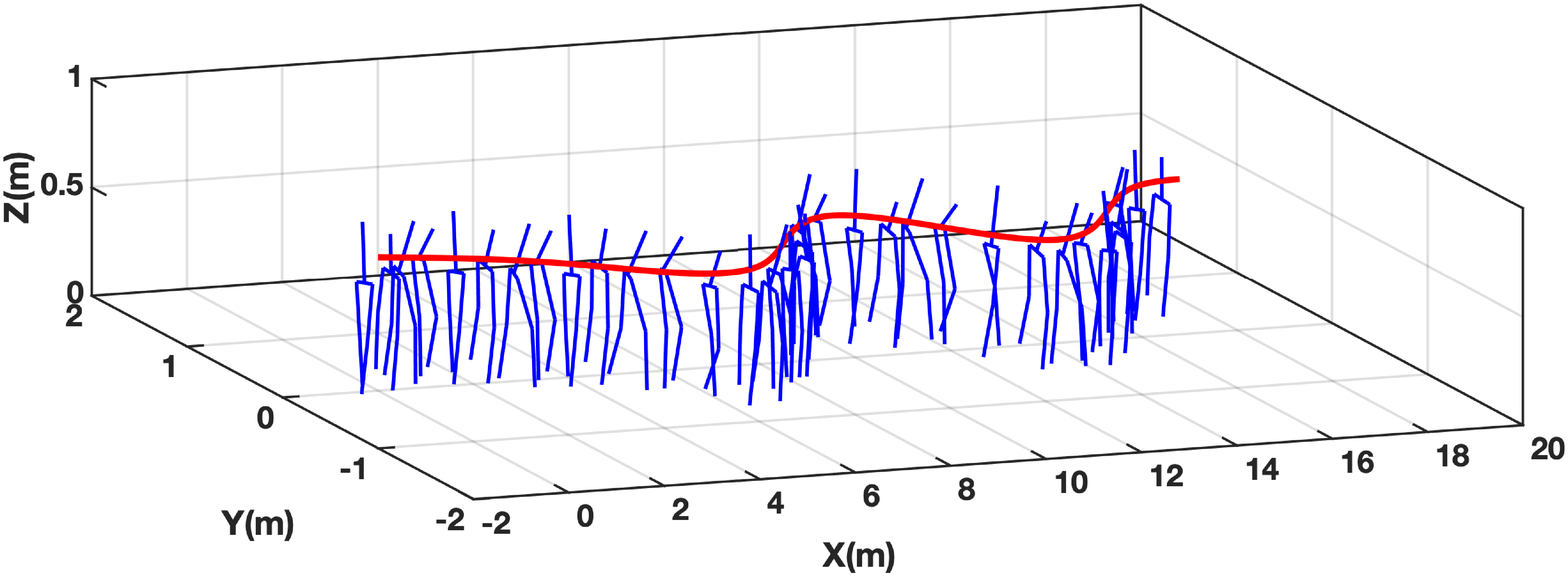}
     \end{subfigure}
     \begin{subfigure}{\columnwidth}
         \centering
         \includegraphics[trim=135 0 165 30,clip,width=0.99\columnwidth]{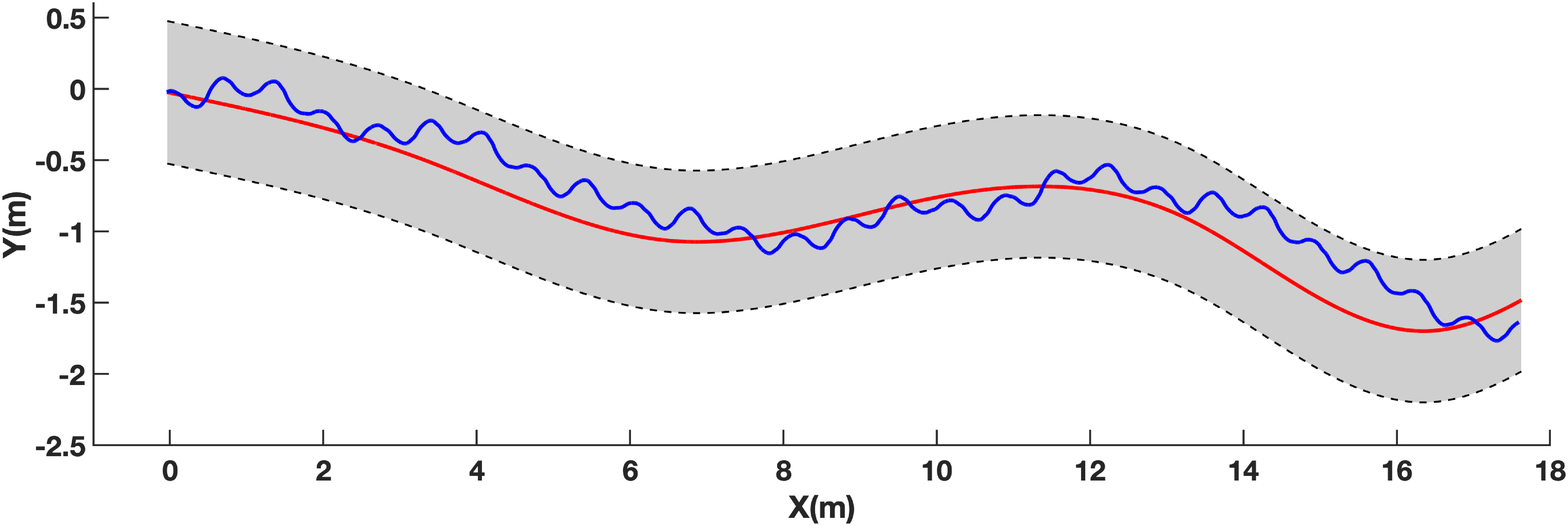}
     \end{subfigure}
    \begin{subfigure}{\columnwidth}
         \centering
         \includegraphics[trim=75 0 80 30,clip,width=0.99\columnwidth]{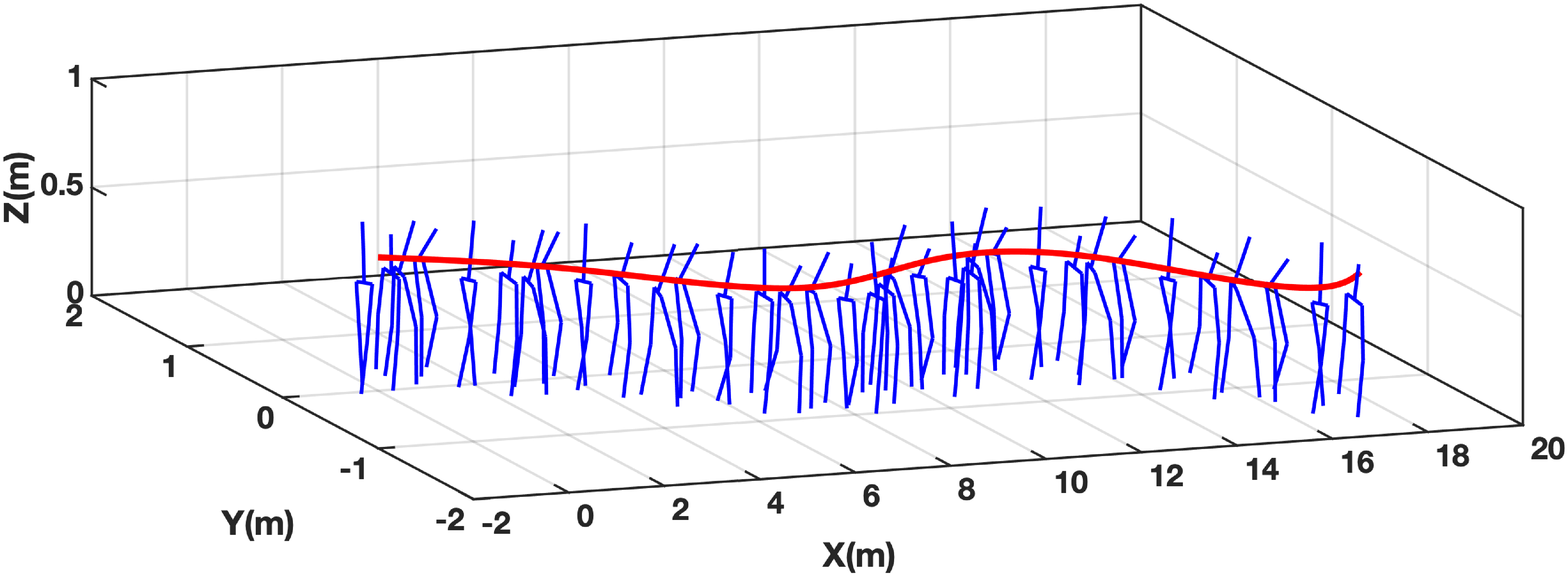}
     \end{subfigure}
     \begin{subfigure}{\columnwidth}
         \centering
         \includegraphics[trim=135 0 165 30,clip,width=0.98\columnwidth]{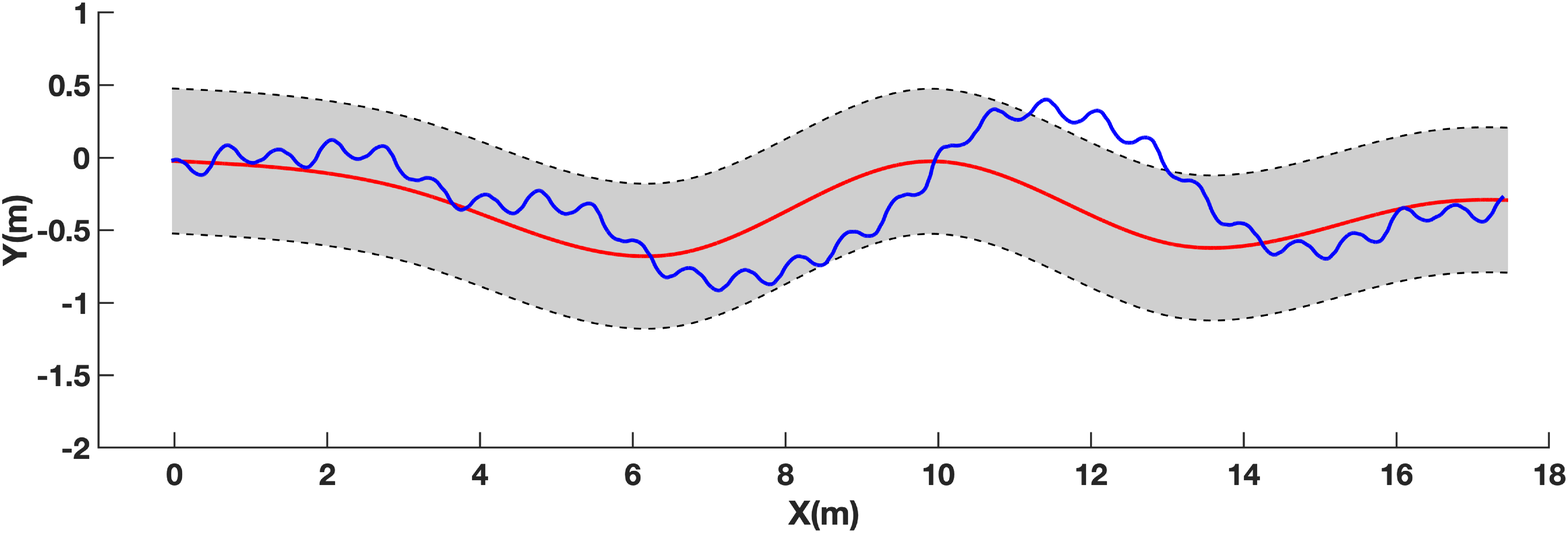}
     \end{subfigure}
    \begin{subfigure}{\columnwidth}
         \centering
         \includegraphics[trim=75 0 80 30,clip,width=0.98\columnwidth]{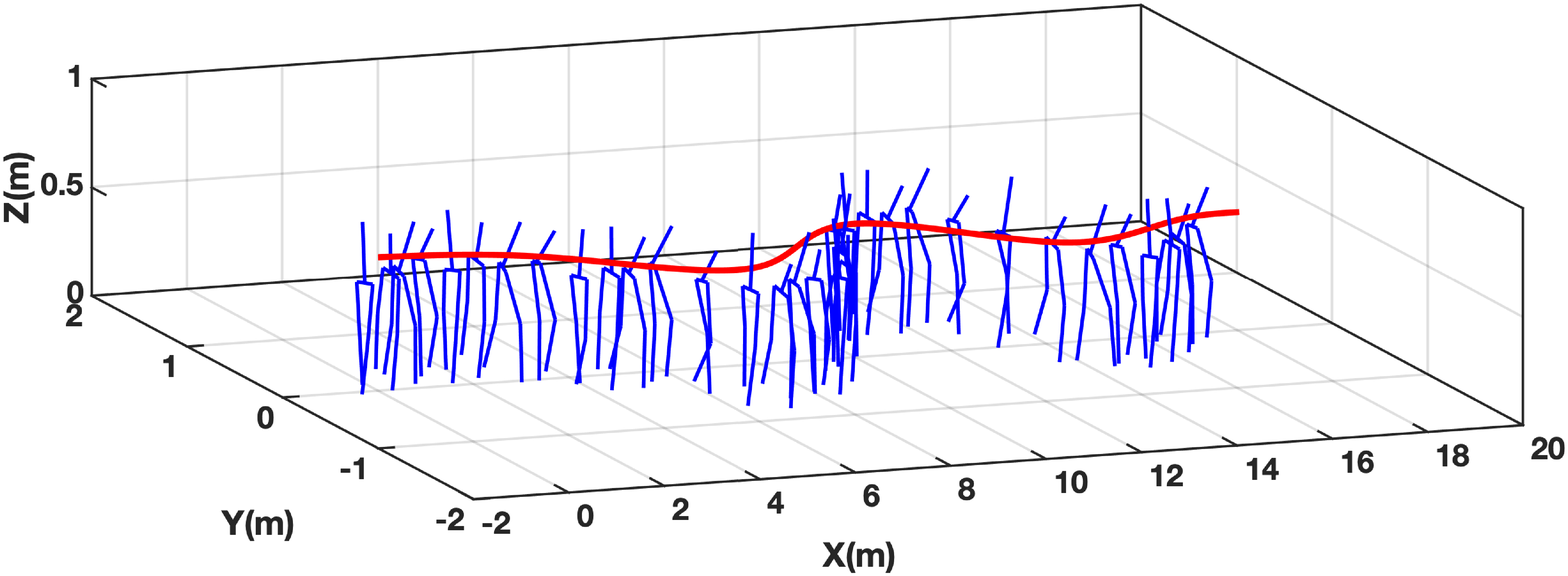}
     \end{subfigure}
        \caption{Tracking performance of the biped under a learned policy. (Left) The intended trajectory $\mathrm{p}_\mathrm{L}(t)$ is in red, the biped's trajectory $\mathrm{p}_\mathrm{R}(t)$ is in blue and the tube of radius $r = 0.5 \mathrm{m}$ around $\mathrm{p}_\mathrm{L}(t)$ is in grey. (Right) Snapshots of the biped for the environments in the left.}
        \label{fig:tubes}
\vspace{-0.2in}
\end{figure*}

Table \ref{table:pac-bayes} also presents estimates of the true cost obtained by simulating the learned policy on 1000 novel environments; that is, environments that were not part of training. It can be seen that with increasing number of environments the PAC-Bayes bounds get closer to the empirical estimate of the true cost. 
To interpret the PAC-bounds presented, consider the last row of the Table \ref{table:pac-bayes}. 
According to Theorem~\ref{thm:PAC-Bayes}, the biped tracks the leader's trajectory while staying in the tube $91.38\%$ of the times with confidence $99\%$. Fig.~\ref{fig:tubes} depicts three examples of applying the learned policy in novel environments.

\vspace{-0.1in}
\begin{table}[h!]
\renewcommand{\arraystretch}{1.3}
\caption{\small PAC-Bayes costs for the interaction task}
\centering
\begin{tabular}{|c||c||c|}
\hline
\bfseries \# Envs  & \bfseries PAC-bound & \bfseries True success \\
(N)&$(1 - C_\mathrm{QPAC}) \times 100$&(estimate)\\
\hline\hline
200 & 82.97\% & 95.57\% \\
\hline
500 & 89.19\% & 95.58\% \\
\hline
1000 & 91.38\% & 95.47\% \\
\hline
\end{tabular}
\label{table:pac-bayes}
\end{table}
\vspace{-0.2in}

\section{Conclusion}
\vspace{-0.05in}

We presented an approach to train a supervisor for gait adaptation for dynamically walking bipedal robots tasked with following a leader's unknown intended trajectory based on interaction. The supervisor is trained to orchestrate switching among a family of gait primitives by minimization of the PAC-Bayes upper bound. This way, the supervisor provides guarantees of generalization, essentially quantifying the risk of deploying a policy to novel leader intentions.
We demonstrated the efficacy of our approach in deriving practical and strong generalization bounds in the case of a dynamic bipedal robot physically collaborating with a leader.

\bibliographystyle{IEEEtran}
\bibliography{bib_lib_ral2021}

\end{document}